\documentclass{article}

\usepackage{float}

\usepackage[preprint]{neurips_2023}

\usepackage[utf8]{inputenc} 
\usepackage[T1]{fontenc}    
\usepackage{hyperref}       
\usepackage{url}            
\usepackage{booktabs}       
\usepackage{amsfonts}       
\usepackage{nicefrac}       
\usepackage{microtype}      
\usepackage{xcolor}         

\title{RoBERTurk: Adjusting RoBERTa for Turkish}

\author{%
  Nuri Tas \\
  Department of Physics\\
  Bogazici University\\
  \texttt{nuri.tas@boun.edu.tr} \\
}

\begin{document}
\maketitle

\begin{abstract}
  We pretrain RoBERTa (\cite{roberta}) on a Turkish corpora using BPE tokenizer. Our model outperforms BERTurk family models  (\cite{berturk}) on the BOUN dataset for the POS task while resulting in underperformance on the IMST dataset for the same task and achieving competitive scores on the Turkish split of the XTREME dataset for the NER task - all while being pretrained on smaller data than its competitors. We release our pretrained model and tokenizer.\footnote{\url{https://huggingface.co/Nuri-Tas/roberturk-base}}
\end{abstract}

\section{Introduction}

Language models such as BERT (\cite{bert}), ELECTRA (\cite{electra}), and RoBERTa (\cite{roberta}) established significant results. However, the careful evaluation of the architectures of these models for the morphologically rich nature of Turkish still needs further investigation. Different characteristics of Turkish, such as flexible word order and agglutinative process, may hinder the performance of contemporary language models, especially in the context of masking algorithms. 

We present a replication of RoBERTa (\cite{roberta}) using Sentencepiece BPE tokenizer (\cite{sentencepiece_bpe}). Our model either outperforms models trained on various Turkish corpora by BERTurk (\cite{berturk}) on the part of speech (POS) tagging task despite being pretrained on a smaller dataset.

\section{Background}
\label{gen_inst}

We briefly review the architecture of RoBERTa (\cite{roberta}) in this section.

The inputs to RoBERTa are tokenized using Byte-Pair Encoding (BPE) (\cite{bpe}) with $50K$ vocabulary size without any preprocessing steps. Tokens are additionally appended with $[BOS]$ and $[EOS]$ special tokens, which denote the beginning and end of a sentence, respectively. Sentences are contiguous text and do not have to be linguistic sentences. 

A random sample of input sequences is then masked with another special token $[MASK]$. Unlike BERT, however, RoBERTa implements a dynamic masking algorithm where new mask patterns are attained at each iteration. Inputs finally go through the transformer model (\cite{transformer}) with $L$ layers using $A$ self-attention
heads and $H$ hidden units without any labels. The pretraining objective is cross-entropy loss of predicting the masked tokens. RoBERTa also removes the next sentence prediction objective during pretraining.

\paragraph{Optimization} RoBERTa uses Adam optimizer (\cite{adam})  with $\epsilon = 1e-6$, $\beta_1 = 0.9$, and $\beta_2 = 0.98$. The learning rate for RoBERTa$_{BASE}$ 
is warmed up to the peak value of $6e-4$ for the first $24K$ updates and linearly decayed to 0. The model is pretrained for maximum $500K$ updates only with sequences of at most $T=512$ length. Note that optimization parameters for fine-tuning on downstream tasks may differ.

\section{Our Setup}
\label{headings}

\paragraph{Implementation}

We use {\fontsize{9}{12}\fontfamily{ptm}\selectfont 
FAIRSEQ%
} (\cite{fairseq}) to pretrain RoBERTa with mixed precision arithmetic.
The model is warmed to the peak value of $1e-5$ for the first $10K$ steps and pretrained for a total of $600K$ steps with mini batches containing 256 samples of maximum length $T=256$.

For tokenization, we use sentencepiece (\cite{sentencepiece_bpe}) library and train BPE on randomly sampled 30M sentences from the training data, which contained around 90M sentences. 

\begin{table}[H]
\caption{Hyperparameters for pretraining RoBERTurk}
\label{hyperparam-pretraining}
    \centering
    \begin{tabular}{lc}
                \toprule
        Hyperparam & RoBERTurk \\
        \midrule
Number of Layers & 12 \\
Hidden size & 1024 \\
FFN inner hidden size & 3072 \\
Attention heads & 12 \\
Attention head size & 64 \\
Dropout & 0.1 \\
Attention Dropout & 0.1 \\
Warmup Steps & 10k \\
Peak Learning Rate & 1e-5 \\
Batch Size & 256 \\
Weight Decay & 0.01 \\
Max Steps & 600k \\
Learning Rate Decay & Linear \\
Adam $\epsilon$ & 1e-6 \\
Adam $\beta_1$ & 0.9 \\
Adam $\beta_2$ & 0.98 \\
        \bottomrule
    \end{tabular}
\end{table}

\subsection{Data}

We make use of two datasets, having a total of 5B tokens and 28GB data size:

• OSCAR (\cite{turkish_oscar}) Turkish split, which is deduplicated.\footnote{The dataset is on  
\href{https://huggingface.co/datasets/oscar/viewer/unshuffled\_deduplicated\_tr}{https://huggingface.co/datasets/oscar/viewer/unshuffled\_deduplicated\_tr}} (27GB).

• The first 12 files of the Turkish split of the processed version of the C4 dataset (\cite{c4_turkish}).\footnote{The dataset is on 
\href{https://huggingface.co/datasets/allenai/c4}{https://huggingface.co/datasets/allenai/c4}}
 Sentences are extracted using the {\fontsize{10}{12}\fontfamily{ptm}\selectfont 
nltk%
} library (\cite{nltk}) in the same preprocessing way by BERTurk (\cite{berturk}). (1GB).

We also note the pretraining data size for the competitor models in Table~\ref{data-sizes}. 

\begin{table}[H]
\caption{Model Pretraining data size}
\label{data-sizes}
    \centering
    \begin{tabular}{lc}
                \toprule
        Model Name & Pretraining data size \\
        \midrule
BERTurk (cased, 128k)    &  35GB \\
BERTurk (cased, 32k)     &  35GB \\
BERTurk (uncased, 128k)  &  242GB \\
BERTurk (uncased, 32k)   &  242GB \\
ConvBERTurk &  35GB \\
ConvBERTurk mC4 (cased)  &  35GB \\
ConvBERTurk mC4 (uncased) &  35GB \\
DistilBERTurk &  35GB \\
ELECTRA BaseException( &  35GB \\
ELECTRA Base mC4 (cased) &  35GB \\
ELECTRA Base mC4 (uncased) &  242GB \\
ELECTRA Small &  242GB \\
RoBERTurk & 28GB \\
        \bottomrule
    \end{tabular}
\end{table}

\subsection{Evaluation}

We finetune the model for part of speech (POS) tagging on BOUN (\cite{boun_dataset}) and IMST (\cite{imst1} and \cite{imst2}) datasets and for named entity recognition (NER) on XTREME (\cite{xtreme}) Turkish split.  BOUN and IMST datasets are annotated in the Universal Dependencies style, and the task is to classify the corresponding label for each word. Similarly, NER task is to classify named entities for each word.

We present the finetuning results in Table~\ref{boun-results}. The results are the average over five runs.  For BOUN and IMST datasets, accuracy is reported, whereas F1 scores are given for the XTREME dataset. While our model outperformed BERTurk family models on BOUN datasets, it yielded less accuracy than its competitors on IMST. Meanwhile, the model achieved competitive scores on the NER task where BERTurk models remarkably accomplished  around 97\%
accuracy.

The hyperparameters for finetuning are given in  Table~\ref{hyperparam-finetuning}. Unlike RoBERTa, the learning rate is kept the same after the warmup process.  

\begin{table}[H]
\caption{Model performance on finetuning tasks}
\label{boun-results}
    \centering
    \begin{tabular}{lccc}
                \toprule
        Model & BOUN & IMST & XTREME\\
        \midrule
        BERTurk (cased, 128k)    &  91.016 ± 0.60  & 96.846 ± 0.42  & 93.8960 ± 0.16 \\
        BERTurk (cased, 32k)     &  91.490 ± 0.10 & 97.096 ± 0.07 & 93.4706 ± 0.09 \\
        BERTurk (uncased, 128k)  &  91.286 ± 0.09 & 97.060 ± 0.07 & 93.4686 ± 0.08 \\
        BERTurk (uncased, 32k)   &  91.544 ± 0.09 & 97.088 ± 0.05 & 92.9086 ± 0.14 \\
        ConvBERTurk &  91.524 ± 0.07 & 97.346 ± 0.07 & \textbf{93.9286 ± 0.07} \\
        ConvBERTurk mC4 (cased)  &  91.724 ± 0.07 & \textbf{97.426 ± 0.03} & 93.6426 ± 0.15 \\
        ConvBERTurk mC4 (uncased) &  91.484 ± 0.12 & 97.338 ± 0.08 & 93.6206 ± 0.13 \\
        DistilBERTurk &  91.044 ± 0.09 & 96.560 ± 0.05 & 91.5966 ± 0.06 \\
        ELECTRA Base &  91.534 ± 0.11 & 97.232 ± 0.09 & 93.4826 ± 0.17 \\
        ELECTRA Base mC4 (cased) &  91.746 ± 0.11 & 97.380 ± 0.05 & 93.6066 ± 0.12 \\
        ELECTRA Base mC4 (uncased) &  91.178 ± 0.15 & 97.210 ± 0.11 & 92.8606 ± 0.36 \\
        ELECTRA Small &  90.850 ± 0.12 & 95.578 ± 0.10 & 90.8306 ± 0.09 \\
        XLM-R (base)     &  91.862 ± 0.16 & 96.492 ± 0.06 & 92.9586 ± 0.14 \\
        mBERT (cased) &  91.492 ± 0.11 & 95.754 ± 0.05 & 93.2306 ± 0.07 \\
        RoBERTurk & \textbf{91.942 ± 0.13}  & 93.812 ± 0.15 & 93.4052 ± 0.16 \\
        \bottomrule
    \end{tabular}
\end{table}

\begin{table}[H]
\caption{Hyperparameters for finetuning RoBERTurk}
\label{hyperparam-finetuning}
    \centering
    \begin{tabular}{lccc}
                \toprule
        Hyperparam & BOUN & IMST & EXTREME \\
        \midrule
Learning Rate & 5e-5 & 1e--5 & 1e-5 \\
Batch Size & 16 & 16 & 16 \\
Weight Decay & 0.06  & 0.06 & 0.1 \\
Max Epochs & 10 &  10 & 10 \\
Warmup Ratio & 0.1 &  0.06 & 0.1 \\
        \bottomrule
    \end{tabular}
\end{table}

\section{Conclusion}

We pretrain RoBERTa on a Turkish corpus and evaluate our model on three different datasets against BERTurk models. Despite being trained on a larger dataset than its competitors, our model achieves competitive scores on BOUN and XTREME datasets but underperforms on IMST.

We release the pretrained model and tokenizer at \url{https://huggingface.co/Nuri-Tas/roberturk-base}.

\section*{Acknowledgements}

We are immensely grateful to Arkadas Ozakin for providing us access to the GPUs at the Kandilli Research Institute and for his helpful advice. We also thank Onur Gungor for pointing out different studies and for his valuable feedback.

\bibliography{references}


\end{document}